\documentclass[letterpaper,11pt,twocolumn]{article}

\newcommand{\reportnumber}{TR-RoboPI-2602}
\newcommand{\reportdate}{November 05, 2026} 
\newcommand{\shorttitle}{BCVS: Budget-aware Correspondence-driven View Selection}
\newcommand{\fulltitle}{Budget-aware Correspondence-driven View Selection for \\ Real-time 3D Perception of Mobile Robots }

\usepackage[letterpaper, top=1in, bottom=1in, left=0.75in, right=0.75in,
            headheight=14pt, headsep=14pt]{geometry}
\usepackage[T1]{fontenc}
\usepackage{lmodern}
\usepackage{microtype}

\usepackage{xspace,xcolor}
\definecolor{UFblue}{HTML}{0021A5}
\definecolor{UForange}{HTML}{FA4616}
\definecolor{lightgray}{gray}{0.95}

\usepackage{cite}
\usepackage{amsmath,amssymb,amsfonts}
\usepackage{amsthm}
\theoremstyle{plain}

\theoremstyle{remark}

\usepackage{graphicx}
\usepackage[size=small, labelfont={bf,sf}, labelsep=period]{caption}
\usepackage{subcaption}
\usepackage{booktabs}
\usepackage{tabularx}
\usepackage{multicol}
\usepackage{multirow}
\usepackage{soul}
\usepackage{siunitx}
\usepackage{physunits}
\usepackage{array}
\usepackage[normalem]{ulem}
\usepackage{makecell}
\usepackage{wrapfig}
\usepackage{lipsum}
\usepackage{ragged2e}

\usepackage[shortlabels]{enumitem}

\usepackage{titlesec}
\titleformat{\section}
  {\normalfont\large\sffamily\bfseries\color{UFblue}}{\thesection}{0.6em}{}
\titleformat{\subsection}
  {\normalfont\normalsize\sffamily\bfseries\color{UFblue}}{\thesubsection}{0.5em}{}
\titleformat{\subsubsection}
  {\normalfont\normalsize\sffamily\itshape}{\thesubsubsection}{0.5em}{}
\titlespacing*{\section}{0pt}{2.0ex plus 0.5ex minus 0.2ex}{1.0ex plus 0.2ex}
\titlespacing*{\subsection}{0pt}{1.5ex plus 0.5ex minus 0.2ex}{0.6ex plus 0.2ex}
\titlespacing*{\subsubsection}{0pt}{1.2ex plus 0.3ex minus 0.2ex}{0.4ex}

\usepackage{fancyhdr}
\makeatletter
\let\NAT@parse\undefined
\makeatother
\usepackage[colorlinks=true, linkcolor=UFblue, citecolor=UFblue,
            urlcolor=UForange]{hyperref}
\makeatletter
\DeclareRobustCommand\onedot{\futurelet\@let@token\@onedot}
\def\@onedot{\ifx\@let@token.\else.\null\fi\xspace}
 
\def\ie{\emph{i.e}\onedot}

\makeatother
\long\def\invis#1{}

\usepackage{gensymb}
\usepackage{algpseudocode}
\usepackage[ruled,linesnumbered,vlined,noend]{algorithm2e}
\long\def\invis#1{}

\hypersetup{
  pdftitle={\fulltitle},
  pdfauthor={Vaishnav Ramesh, Shivali J. Patel, Md Jahidul Islam},
  pdfsubject={\reportnumber}
}

\begin{document}


\begin{titlepage}
\centering
\sffamily

{\color{UFblue}\rule{\textwidth}{2pt}}\par\vspace{4pt}
\begin{minipage}{0.5\textwidth}
  \raggedright
  {\large\bfseries\color{UFblue} TECHNICAL REPORT}
\end{minipage}%
\begin{minipage}{0.5\textwidth}
  \raggedleft
  {\small\bfseries Report No.\ \reportnumber}\\[2pt]
  {\small \reportdate}
\end{minipage}\par
\vspace{4pt}{\color{UFblue}\rule{\textwidth}{0.6pt}}\par

\vspace{1.4in}

{\Large\bfseries\color{UFblue} \fulltitle \par}

\vspace{0.3in}
{\color{UForange}\rule{0.25\textwidth}{1.5pt}}\par
\vspace{0.3in}

{\large
Vaishnav Ramesh$^1$, Jacob Carter$^2$, Shivali J. Patel$^1$, Sanjeev J. Koppal$^2$, and Md Jahidul Islam$^1$ \par}
\vspace{0.15in}
{\normalsize\color{gray}
$^1$RoboPI Laboratory, ECE Department, University of Florida \\ 
$^2$FOCUS Laboratory, ECE Department, University of Florida \\ \par}

\vspace{0.6in}

\begin{minipage}{0.88\textwidth}
\colorbox{lightgray}{%
\begin{minipage}{\dimexpr\textwidth-2\fboxsep\relax}
\vspace{6pt}
\rmfamily\small
{\sffamily\bfseries\color{UFblue} Abstract}\par\vspace{4pt}
\justifying
3D visual perception and scene reconstruction are critical for autonomous robots and intelligent transportation, yet real-world deployments often generate long, redundant image sequences that hinder downstream processing. This report investigates BCVS (budget-aware correspondence-driven view sampling), a pose-free view selection algorithm that offers 22+ FPS on a CPU, enabling online filtering for efficient 3D perception by mobile robots. Unlike existing approaches that rely on full structure-from-motion (SfM) pipelines or camera pose estimation, BCVS employs a purely correspondence-based scoring strategy that prioritizes view novelty while maintaining sufficient overlap for geometric stability. Given a user-specified budget, BCVS iteratively selects the most distinct frames and suppresses near-duplicates, producing compact subsets that preserve connectivity and minimize reconstruction loss. Extensive experiments on benchmark show that BCVS outperforms CPU-friendly baselines at tight budgets and remains competitive as budgets scale. Field trials with large aerial and marine robotics data further validate its practicality and scalability: BCVS reduces redundancy up to 16\% while sustaining robust SfM registration and high-quality 3D reconstructions.

\vspace{6pt}
{\sffamily\bfseries\color{UFblue} Keywords:}
3D reconstruction; view selection; image correspondence; real-time vision.
\vspace{6pt}
\end{minipage}}
\end{minipage}

\vfill

{\color{UFblue}\rule{\textwidth}{0.6pt}}\par\vspace{4pt}
{\footnotesize\color{gray} Authors are with the University of Florida (UF), Gainesville, FL, USA. Dr.~Sanjeev~J.~Koppal holds concurrent appointments as an Associate Professor at ECE-UF and a Scholar at Amazon Robotics. This report's activities are performed independently at UF. }
\end{titlepage}


\begin{figure*}[t]
     \centering
     \includegraphics[width=0.96\linewidth]{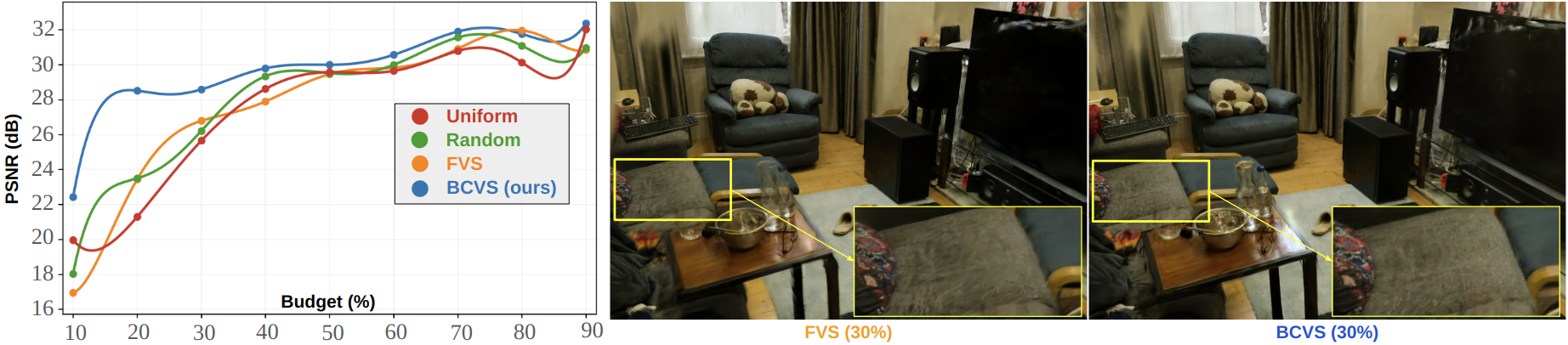}%
     \vspace{-1.5mm}
     \caption{Performance margins for view selection strategies are shown with \textit{budget vs. reconstruction quality} scores on MipsNerf-360 benchmark~\cite{barron2022mip}. \textbf{Top:} PSNR as a function of view budget for four selectors -- Uniform, Random, FVS, and the proposed BCVS algorithm. BCVS achieves higher PSNR across most budgets compared to the current state-of-the-art FVS~\cite{xiao2024nerf}, with the largest gains at low budgets, where view choice is particularly critical. \textbf{Bottom:} A sample rendering comparison at $30$\% budget with insets highlighting texture on the sofa and table region; as seen, FVS exhibits blur and residual artifacts, whereas BCVS preserves sharper detail and cleaner boundaries.}%
     \vspace{-5mm}
     \label{fig:BCVS_comparison}
 \end{figure*}
 
\section{Introduction}

Fast and accurate 3D reconstruction is critical for mobile robots operating in ground, aerial, and marine domains, supporting tasks such as dense mapping, long-horizon navigation, mission planning, and downstream scene understanding, including infrastructure inspection, asset monitoring, and structural assessment~\cite{selvaratnam20253d,shan2020lio}. 
Contemporary visual pipelines range from geometric approaches such as Structure from Motion (SfM) and SLAM~\cite{ullman1979interpretation} to learned neural scene representations~\cite{mildenhall2021nerf,kerbl20233d}, which transform onboard multi-view imagery into metrically consistent 3D structure and radiance fields, offering rich geometric and appearance cues for perception systems. In practice, however, mobile robots often depend on SLAM for real-time~autonomy~\cite{rahman2019svin2} while relegating high-fidelity reconstruction to post-processing due to constraints in power, memory, and latency.


Real-world deployments often generate long, redundant mission data that substantially increases the computational cost of image matching, bundle adjustment (BA), and downstream training. In SfM pipelines, computation scales with the number of input views: retrieval methods are used to prune the quadratic complexity of pairwise matching, while BA repeatedly solves large-scale optimization problems. As a result, the sheer data volume becomes a primary bottleneck~\cite{vocabtree,schoenberger2016sfm}. Similarly, neural 3D reconstruction methods scale with the number of images and sampled rays, such that retaining fewer but more informative views can both accelerate training and improve numerical conditioning~\cite{mildenhall2021nerf,kerbl20233d}. These challenges motivate \textit{budget-aware view selection} strategies that compress missions by preserving only distinctive yet geometrically informative frames. 

Existing approaches to view selection generally fall into two categories: next-best-view (NBV) planning and offline subset selection. NBV planners actively guide data acquisition by selecting the next camera pose according to predicted information gain or uncertainty~\cite{xiao2024nerf,jin2023neu,pan2022activenerf}. While these methods can achieve high-fidelity reconstructions, they require expensive model-in-the-loop updates at each step, which limits scalability. In contrast, offline subset-selection methods only operate post-mission, \ie, after full data collection. Redundancy-aware strategies~\cite{wang2025diversitydrivenviewsubsetselection} leverage images with pre-estimated camera poses, whereas coverage-based approaches, such as NeRF Director~\cite{xiao2024nerf}, rely on coarse SfM outputs to enforce spatially uniform sampling. Although effective at reducing redundancy, these offline methods remain dependent on SfM-derived geometry or camera pose information, constraining their applicability in settings where such priors are unavailable or computationally infeasible.
    
In this report, we present \textbf{Budget-aware Correspondence-driven View Sampling (BCVS)} -- a fast, camera pose–free algorithm that selects views by \textit{regularizing} novelty (information variance) in views from keypoint correspondences alone. Using only images as input, BCVS uses a lightweight feature matching step to estimate inter-frame motion and select diverse viewpoints, without relying on external poses or point clouds. Given a computational budget as prior, the \textit{view candidates} are favored when they add new visual information while retaining enough overlap to remain connected. 
Operating in a short temporal neighborhood, BCVS incrementally prefers informative change, suppresses redundancy, and then lightly adjusts the set to meet a target budget (in maximum number of views). This budget-aware process produces a compact, viewpoint-diverse subset that can be utilized for fast 3D reconstruction with minimal loss.

Notably, our focus is on real-time view selection for 3D robot perception with onboard CPU processing constraints. A fast view selection operation is crucial to keep pace with the camera stream, while the bulk of the processing power is typically dedicated to perception and control systems~\cite{realtimeplenoptickatherine2016}. 
However, most existing approaches are mainly offline, requiring a trained model, large embedding encoders, and/or a full SfM pipeline with accurate camera poses -- which incur latency and memory overhead~\cite{sun2021learning,kaushal2019learning}. To close this gap, we adopt a \textbf{pose-free and CPU-only design} that relies on lightweight keypoint correspondences to admit distinct views while preserving overlap for stable reconstruction. In practice, our system processes frames at \textit{$22.3$} FPS on CPU, enabling online filtering close to the camera rate ($30$~FPS). A fast budget-adjustment pass then consolidates the stream into a compact subset for downstream 3D reconstruction. This hybrid workflow delivers the benefits of an efficient, on-board selection without requiring deep visual inference or precomputed camera poses as prior.

We evaluate BCVS on the MipNeRF-360 benchmark~\cite{barron2022mip}, and further assess it on marine and aerial robotics trial data where slow platform motion yields long stretches of redundant frames. For fairness and to match our deployment constraints, we compare against real-time CPU-based approaches available in the literature: Random, Uniform, and pose-based Farthest View Sampling (FVS)~\cite{xiao2024nerf}; as Fig~\ref{fig:BCVS_comparison} demonstrates, BCVS achieves the best performance under tight computational budgets. We also demonstrate its real-time utility for \textbf{CPU-based online filtering} of large (long) missions, without which, 3D reconstruction onboard mobile robots is computationally infeasible. 


\vspace{1mm}
\noindent
This report makes the following contributions:
\begin{enumerate}[label={\arabic*}.,nolistsep,leftmargin=*] 
\item We introduce BCVS, which selects views by regularizing novelty with correspondence overlap using lightweight image features alone, without requiring a full-scale SfM pipeline or camera poses as prior. BCVS employs a \textit{budget-aware} selection procedure to generate compact subsets that preserve coverage and geometric consistency, yielding \textit{minimal reconstruction loss} downstream.
\item BCVS is computationally efficient, offering \textbf{22+ FPS} on CPU and includes a fast budget-adjustment pass that consolidates the stream into a compact subset, enabling timely online filtering onboard mobile robots without requiring pretrained embeddings or pose estimation.
\item Comprehensive evaluations on \textit{MipNeRF-360} and field experimental data confirm the effectiveness of BCVS. It consistently matches or surpasses pose-based selection methods and clearly outperforms all baselines at low budgets, while remaining competitive as budgets increase.
\end{enumerate}

\begin{figure*}[t]
     \centering
     \includegraphics[width=\textwidth]{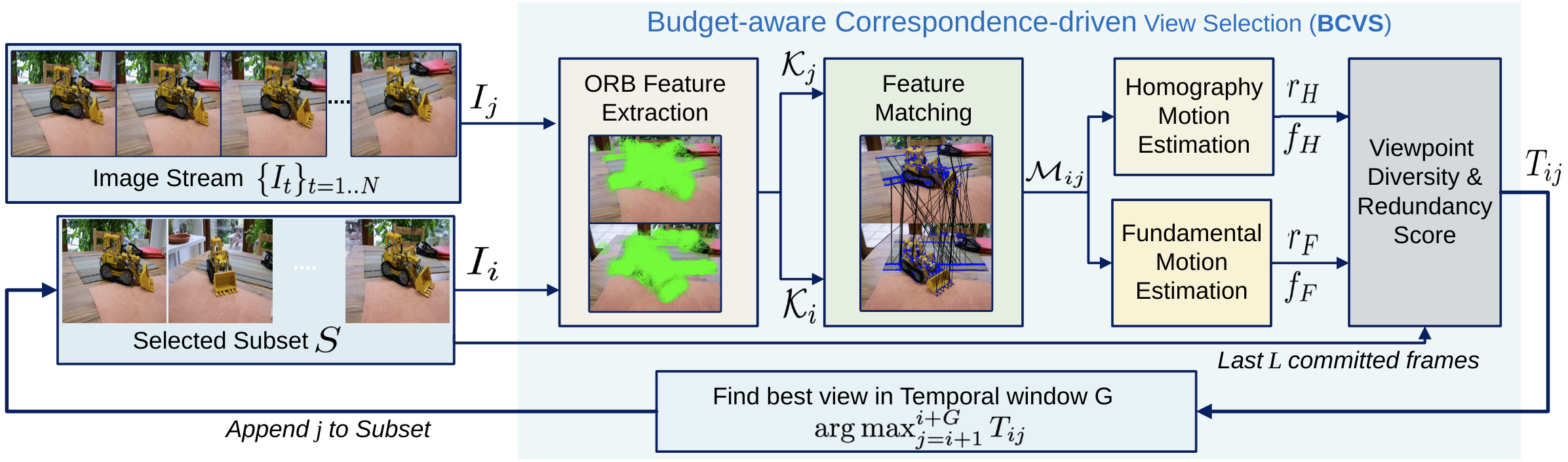}%
     \vspace{-1mm}
     \caption{An overview of the proposed BCVS filtering pipeline for view selection is illustrated. First, ORB features are extracted and matched between frames $I_i$ and $I_j$; from these matches, we estimate a homography $H_{ij}$ and a fundamental matrix $F_{ij}$ to obtain inlier ratios and normalized flows $\big(r_H,f_H\big), \big(r_F,f_F\big)$. These statistics drive the viewpoint diversity and redundancy terms that compose the pairwise scoring function $T_{ij}$. Over a short commit window $G$, we select the top-scoring candidate \ie, the most informative, distinct viewpoint with sufficient overlap and append it to the subset $S$.}%
     \label{fig:Pipeline}
 \end{figure*}

\section{Background \& Related Work}\label{sec:background}
\vspace{-1mm}
Modern reconstruction pipelines encompass a range of methods, including geometric, hybrid, and neural approaches. SfM and SLAM pipelines recover camera trajectories and sparse 3D structure from multi-view imagery~\cite{schoenberger2016sfm,frahm2010building,mur2015orb,placed2023surveyactivesimultaneouslocalization}. 
These systems combine robust feature matching, incremental mapping, and bundle adjustment for metrically consistent models suitable in downstream planning and localization. Neural scene representations~\cite{mildenhall2021nerf} learn a continuous radiance field that renders high-fidelity novel views, trading heavy training for photorealism and smoothness. In particular, 3D Gaussian Splatting~\cite{kerbl20233d} offers a fast, explicit alternative: it optimizes a set of anisotropic Gaussians with per-primitive appearance, enabling real-time rendering while preserving rich geometry/appearance cues.

Moreover, active NBV (next-best-view) planners choose the next \textit{camera view} by maximizing expected information gain or uncertainty reduction under a continually updated scene model. ActiveNeRF~\cite{pan2022activenerf} estimates per-ray uncertainty to greedily reduce it. For instance, FisherRF~\cite{jiang2024fisherrfactiveviewselection} models the \textit{Fisher information} of NeRF parameters and selects viewpoints by maximizing the expected information gain, whereas BavsNeRF methods~\cite{li2024bavsnerf} select batches of complementary views from a partially trained field. On the other hand, IOVS4NeRF models~\cite{xie2025iovs4nerf} target large-scale settings with uncertainty-guided incremental selection.

Offline methods operate after the full trajectory has been collected to prune near-duplicates while preserving geometric/scene coverage. NeRF Director~\cite{xiao2024nerf} evaluates \textit{training view} selection for neural volume rendering and shows which frames materially impact the quality and convergence of the reconstruction. It introduces Farthest View Sampling (FVS) and Information-Gain Sampling (IGS) to enforce uniform, diverse coverage without retraining. Besides, redundancy-aware camera selection~\cite{wang2025diversitydrivenviewsubsetselection} works on images with pre-estimated camera poses, scoring candidates by overlap or diversity to retain a compact subset that maintains coverage while minimizing viewpoint redundancy. Lastly, ConMax3D~\cite{malhotra598conmax3d} selects frames without requiring camera poses; it maximizes the concept or semantic diversity for filtering views based on image embeddings. 

Despite considerable progress in recent years, several view-selection methods still rely on SfM outputs or pre-computed camera poses, which are not always available in real-time robotics applications. Besides, NBV approaches require a model-in-the-loop or high-dimensional image embeddings. We address these limitations with a purely feature-matching-based method that eliminates the need for camera poses, enabling real-time use by mobile robots.

\newpage
\section{Methodology}
\subsection{Overall Computational Pipeline}
We introduce BCVS, a fast view-selection algorithm that produces a compact, viewpoint-diverse subset $S$ from an image stream $\{I_t\}_{t=1..N}$ under a target budget $K$ for downstream 3D reconstruction. We construct the subset $S$ in a single pass: for each incoming frame $I_j$, we compare it to the last committed view and to what is already represented in $S$, to compute a score $T_{ij}$ that rewards \textit{informative viewpoint change} while penalizing near-duplicates, visually static, or redundant views. Using a short temporal window $G$, we retain a candidate with the highest score and commit it when the window closes; $S$ thus grows by repeatedly adding the next most informative, least redundant frame. Fig.~\ref{fig:Pipeline} depicts the selection pipeline in detail.

After one pass of processing, we further adjust the view selection if the budget is under-utilized or overshot. If $|S|<K$, we densify by inserting frames into the largest temporal gaps in $S$ to improve coverage and connectivity between neighbors. On the other hand, if $|S|>K$, we prune by removing frames with the smallest marginal contribution to viewpoint diversity relative to their neighbors. The resulting subset $ S~( |S|\approx K)$ preserves geometric diversity and connectivity, enabling scene reconstruction with minimal degradation compared to using all frames.

\subsection{Feature Matching and Scoring}
Unlike prior methods~\cite{xiao2024nerf,wang2025diversitydrivenviewsubsetselection} that rely on camera poses, we estimate viewpoint redundancy and diversity using only images via feature matching. For each pair $(i,j)$ in $S$, we detect ORB~\cite{ORBfeatures} keypoints and binary descriptors; ORB offers repeatability under moderate viewpoint/illumination change and enables fast matching with a small memory footprint. We form putative correspondences using a BF–Hamming matcher and retain good matches via Lowe's ratio test~\cite{lowe2004distinctive}. Let $\mathcal{M}_{ij}$ be the resulting set of good correspondences between $I_i$ and $I_j$, and let $\mathcal{K}_i, \mathcal{K}_j$ be their detected keypoint sets. We define the \textit{match reliability} ratio as:
\begin{equation}
   {mr}_{ij} \;=\; {|\mathcal{M}_{ij}|} :{\min\!\left(|\mathcal{K}_i|,\,|\mathcal{K}_j|\right)}.
   \label{eq_ratio}
\end{equation}
This pose-free correspondence layer supplies reliable point pairs at low cost, enabling a fast selection pipeline.

For pair $(i,j)$, whether $I_j$ contributes a \textit{new viewpoint} -- is assessed with a per-pair score $T_{ij}$; it is composed of three terms: (i) a \textit{viewpoint diversity} score $D_{ij}$ that measures how much the camera viewpoint has changed, using geometry-consistent motion between the images; (ii) a \textit{redundancy penalty} $R_{ij}$ that captures how visually similar and motion-wise static the pair is; and (iii) a \textit{global redundancy} $R_{ij}^G$ that penalizes frames that are similar to selected views from $S$. 

Given $\mathcal{M}_{ij}$, a homography $H_{ij}$ and a fundamental matrix $F_{ij}$ are estimated; $H_{ij}$ captures planar and pure rotational motion, while $F_{ij}$ encodes the epipolar geometry, capturing translation-induced parallax. In our estimation, each fit returns an inlier mask, $\mathcal{I}_H \subset \mathcal{M}_{ij}$ and $\mathcal{I}_F \subset \mathcal{M}_{ij}$, consisting of matches within respective models under a reprojection threshold. The corresponding inlier ratios $r_H$ and $r_F$ are $r_H={|\mathcal{I}_H|}:{|\mathcal{M}_{ij}|}$ and $r_F={|\mathcal{I}_F|}:{|\mathcal{M}_{ij}|}$, respectively.

With this, we define the inlier flow parameters $(f_H,f_F)$ to quantify the motion under each geometry. It is computed by mapping the median 2D displacement of the inlier correspondences to $[0,1]$. Let $\bar d$ denote the image diagonal, and $\lambda > 0$ a flow-scale constant. then,
\begin{equation*}
\begin{aligned}
\footnotesize
f_H &= \tanh\!\left(\frac{ \operatorname{median}_{(x,x')\in\mathcal{I}_H}\!\|x'-x\|_2}{\lambda\,\bar d}\right), \\
f_F &= \tanh\!\left(\frac{\operatorname{median}_{(x,x')\in\mathcal{I}_F}\!\|x'-x\|_2}{\lambda\,\bar d}\right). \\
\end{aligned}
\end{equation*}
Effectively, the pairs $(r_H,f_H)$ and $(r_F,f_F)$ summarize, respectively, the rotation/planar motion and translation constrained by epipolar geometry.

\vspace{1mm}
\noindent 
\noindent 
\textbf{Viewpoint Diversity Score}:We factor diversity into two complementary terms: a \emph{connectivity} term $C_{ij}$ that measures reliable overlap between the images, and a \emph{parallax} term $P_{ij}$ that rewards translation-induced parallax.
\begin{equation}
\begin{aligned}
C_{ij} &= \mathrm{mr}_{ij}\, r_F \\ 
P_{ij} &= (r_F\, f_F)^{q}, \quad q\in(0,1],\\
D_{ij} &= w_{\mathrm{conn}}\, C_{ij} + w_{\mathrm{par}}\, P_{ij}.
\end{aligned}
\label{eq:Dij}
\end{equation}
This design \emph{embodies novelty regularized by overlap}: the connectivity term ensures that new views are supported by sufficient, geometry-consistent correspondences (overlap), while the parallax term promotes genuine baseline change (novelty). Thus, choosing frames with high $D_{ij}$ prioritizes candidates that expand scene coverage over frames that repeat already observed content.

\vspace{1mm}
\noindent 
\textbf{Viewpoint Redundancy Penalty}: $R_{ij}$ is measured using the structural similarity index (SSIM)~\cite{wang2004image}, combined with geometry cues to reflect limited motion, and is denoted as
\begin{equation}
  R_{ij} \;=\; s_{ij}\,\Big[r_H\,(1-f_H)\Big]^r
  \label{eq_Rij}
\end{equation}
where $r\in[0,1]$ and $s_{ij}$ is the SSIM between the downsampled versions of $I_i$ and $I_j$. The penalty increases when the pair is geometry-consistent (large $r_H$) yet shows limited motion (small $f_H$), and the images are highly similar. 

\vspace{1mm}
\noindent 
\textbf{Global Redundancy Penalty}: $R^{\mathrm{G}}_{j}$ suppresses selections that are redundant with those already in $S$; that is, the current candidate $I_j$ is compared against the last $L$ committed frames in $S$. Let $S_L \subseteq S$ denote the indices of the most recent $L$ selections. This penalty applies a \textit{hinge loss} as:
\begin{equation}
  R^{\mathrm{G}}_{j}
=\max\,\Bigl(0,\ (\max_{s\in S_L} c_{js}-\tau)\Bigr), \, \tau\in(0,1).
  \label{eq_RG}
\end{equation}

\begin{figure}[t]
    \centering
    \includegraphics[width=0.98\columnwidth]{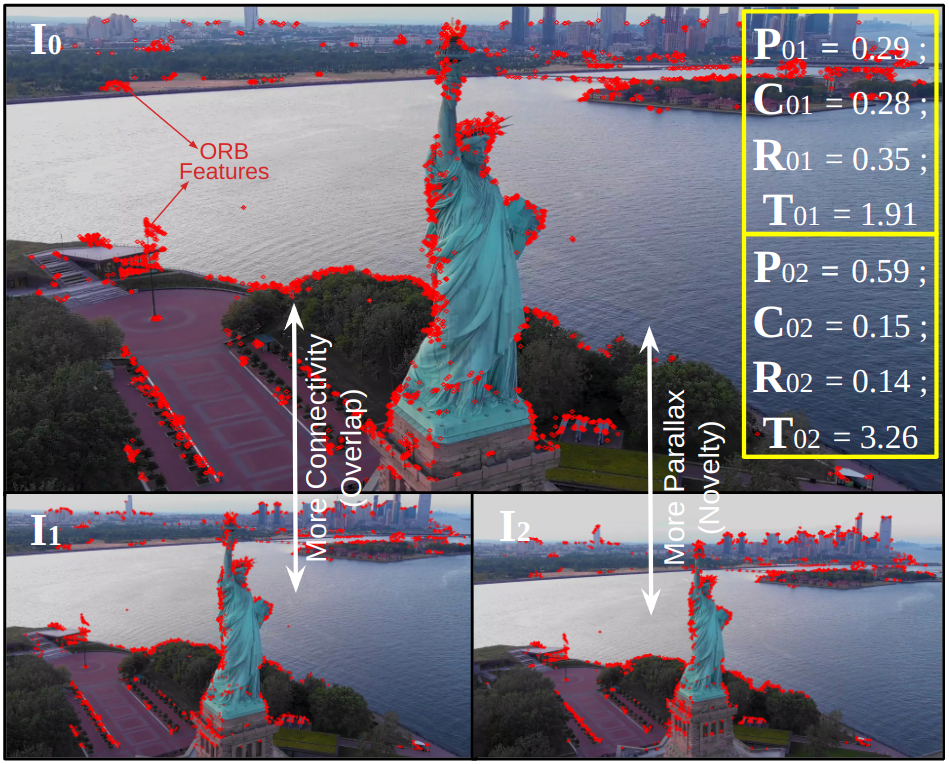}%
    \vspace{-1mm}
    \caption{Overview of our scoring function $T_{ij}$: given an anchor frame $I_{0}$, we compare candidates $I_{1}$ and $I_{2}$; $I_{1}$ offers higher overlap/connectivity, while $I_{2}$ provides greater parallax/novelty. BCVS balances parallax and overlap to keep views informative yet connected; in this example, the stronger parallax of $I_{2}$, outweighs the extra overlap of $I_{1}$, thus $I_{2}$ is preferred for selection.} 
    \label{Fig:ScoringFunction}
    \vspace{-3mm}
\end{figure}

%
\noindent
\textbf{Total Score}: $T_{ij}$, the total pair-to-pair score is then given by
\begin{equation}
T_{ij} = w_{\text{div}}\,D_{ij} - w_{\text{red}}\,R_{ij}-w_{\text{glb}}\,R^{G}_{j}
\label{eq_total}
\end{equation}
where $w_{\odot}$s are hyperparameters weighting the viewpoint diversity, redundancy penalty, and global redundancy scores. Fig~\ref{Fig:ScoringFunction} illustrates how we select images using $T_{ij}$, balancing novelty against overlap to choose distinct views.

\begin{algorithm}[t]
\caption{Budget-aware correspondence-driven view sampling (BCVS).}
\label{Algorithm:Selection_pipleine}
\KwIn{Stream $\{I_t\}_{t=1}^N$, budget $K$, window $G$}
\KwOut{Selected subset $S$}
$S \gets \{1\}$; $i \gets 1$; $b \gets \varnothing$; $v \gets -\infty$\;
\For{$j \gets 2$ \KwTo $N$}{
  Compute ORB correspondence $\mathcal{M}_{ij}$\; 
  Compute match ratio $\mathrm{mr}_{ij}$\;
  \If{$|\mathcal{M}_{ij}| \ge n_{\min}$}{
    Estimate $H_{ij}, F_{ij}$;
    Compute inlier ratios $r_H, r_F$, and flow parameters $f_H, f_F$;
  }
  \Else{
    $r_H, r_F, f_H, f_F \gets 0$\;
  }
  Compute downsampled SSIM $s_{ij}$\;
  Compute diversity score $D_{ij}$\;
  Compute redundancy penalties $R_{ij}, R_{ij}^G$\;
  Compute total score $T_{ij}$\;
  Update best index and score $(b,v)$\;
  \If{$j - i \ge G$}{
    Commit $b$ to $S$; $i \gets b$; $(b,v) \gets (\varnothing,-\infty)$\;
  }
}
\While{$|S| < K$}{
  Find longest gap $(L,R)$ in $S$\;
  Insert index $j^\star$ maximizing $J_{L,j,R}$\;
}
\While{$|S| > K$}{
  Remove index $t$ with the least $\Gamma_t$\;
}
\Return{$S$}\;
\end{algorithm}

\subsection{Algorithm and Implementation Details}
The proposed budget-aware correspondence-driven view sampling (BCVS) pipeline is summarized in 
Algorithm~\ref{Algorithm:Selection_pipleine}. It performs fast view selection with a fixed temporal window of size $G$ to maintain temporal consistency. First, it initializes the selected set $S$ with the first frame; then, for each window of candidates that follows, it evaluates the per-pair score $T_{ij}$ based on Eq.~\ref{eq_total}. The frame with the highest score is added to $S$ as the next view, the window advances from this new anchor, and the process iterates. Using a bounded temporal neighborhood discourages abrupt jumps, preserves continuity in motion and content, and steadily accumulates diverse viewpoints until the entire stream is processed.

\begin{table*}[t]
\centering
\caption{Quantitative results for reconstruction quality given a selection budget on \textit{MipNeRF-360} dataset~\cite{barron2022mip}. Each cell reports PSNR/SSIM scores across scenes on the held-out split, using the same 3DGS training protocol for all methods; only the selected training views differ. Budgets indicate the \% of views retained; scores are shown as \textit{mean} $\pm$ \textit{standard deviation}; best scores are in \textbf{bold}.}
\setlength{\aboverulesep}{0pt}
\setlength{\belowrulesep}{0pt}
\renewcommand{\arraystretch}{1.1}
\resizebox{\textwidth}{!}{
\begin{tabular}{l||c|c|c|c|c|c|c|c}
\toprule
 & \multicolumn{8}{c}{Percentage of \textbf{Views} Selected (\%)} \\
\cline{2-9}
\textbf{Sampler}~($\downarrow$) & 100 & 90 & 70 & 50 & 40 & 30 & 20 & 10 \\
\midrule
\multicolumn{9}{c}{\textit{PSNR}} \\
\midrule
Random & \multirow{4}{*}{$\mathbf{29.40\pm2.57}$} & $28.11\pm4.52$ & $27.66\pm4.76$  & $24.69\pm4.89$ & $25.41\pm4.72$ & $23.58\pm4.11$ & $21.78\pm2.82$ & $18.46\pm3.17$ \\
Uniform & & $28.11\pm4.60$ & $27.46\pm4.95$ & $\mathbf{26.12\pm4.06}$ & $25.37\pm4.09$ & $22.09\pm3.39$ & $20.79\pm3.59$ & $17.65\pm2.29$ \\
FVS~\cite{xiao2024nerf} &  & $28.24\pm2.45$ & $\mathbf{28.03\pm3.18}$ & $26.06\pm3.77$ & $25.49\pm3.89$ & $23.62\pm3.66$ & $22.37\pm0.78$ & $16.79\pm1.29$ \\
\textbf{BCVS~(Ours)} &  & $\mathbf{28.43\pm3.93}$ & $27.87\pm4.36$ & $26.10\pm4.94$ & $\mathbf{26.08\pm4.11}$ & $\mathbf{24.44\pm3.49}$ & $\mathbf{24.85\pm2.92}$ & $\mathbf{20.60\pm1.33}$ \\
\midrule
\multicolumn{9}{c}{\textit{SSIM}} \\
\midrule
Random & \multirow{4}{*}{$\mathbf{0.885\pm0.06}$}& $0.852\pm0.122$  & $0.828\pm0.158$ & $0.743\pm0.196$ & $0.778\pm0.203$ & $0.722\pm0.168$ & $0.730\pm0.085$ & $0.692\pm0.105$ \\
Uniform & & $0.843\pm0.141$ & $0.825\pm0.171$ & $\mathbf{0.819\pm0.133}$ & $0.771\pm0.193$ & $0.673\pm0.205$ & $0.687\pm0.153$ & $0.573\pm0.133$ \\
FVS~\cite{xiao2024nerf} &  & $0.858\pm0.105$  & $\mathbf{0.851\pm0.117}$ & $ 0.806 \pm 0.152$ & $ \mathbf{0.808 \pm 0.155}$ & $0.713 \pm 0.222$ & $0.776\pm0.048$ & $0.528\pm0.083$ \\
\textbf{BCVS~(Ours)} &  & $\mathbf{0.861\pm0.107}$  & $0.846\pm0.126$  & $0.803\pm0.172$ & $0.793\pm0.167$ & $\mathbf{0.756\pm0.171}$ & $\mathbf{0.812\pm0.072}$ & $\mathbf{0.675\pm0.102}$ \\
\bottomrule
\end{tabular}
}
\label{results}
\vspace{-2mm}
\end{table*}
\begin{figure*}[h]
    \centering
    \includegraphics[width=\textwidth]{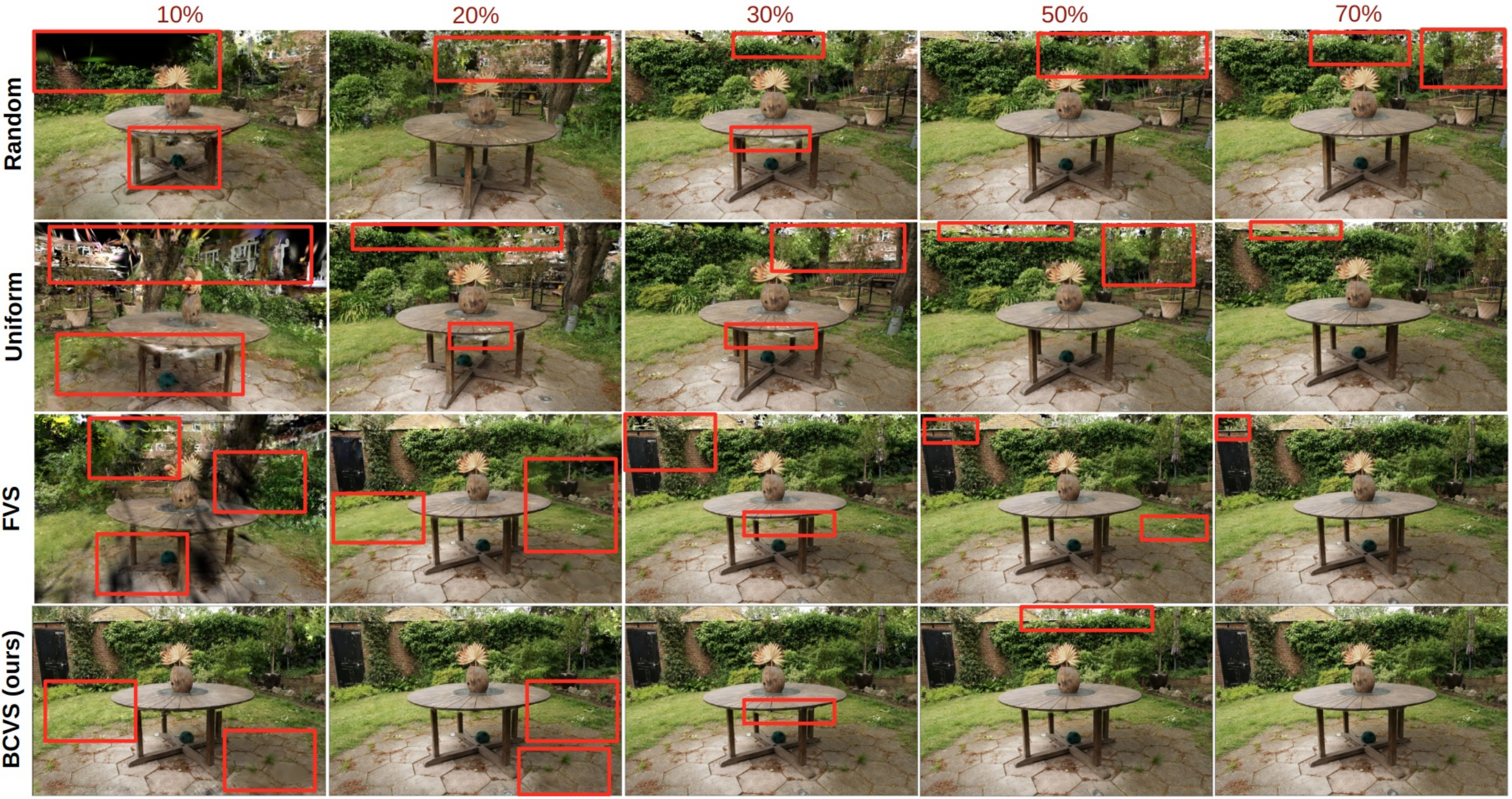}%
    \vspace{-1mm}
    \caption{Qualitative comparison of reconstructed views across sampling budgets ($10$-$70$\%) for \texttt{garden} scene. The red boxes mark major errors and inconsistencies such as missing geometry, distortions, and blurring. Random and Uniform leave large viewpoint gaps, while FVS spreads poses broadly but often with poor overlap, leading to \textit{ghosting} and \textit{unstable} rendering. BCVS balances both novelty and overlap, producing sharper textures and more complete geometry, and generalizes better to unseen views.} 
    \label{Fig:qualitative}
\end{figure*}

After all frames are processed, the selected subset $S$ may not satisfy the target budget $K$. When $|S|<K$, temporal gaps between consecutive selections are examined, and each candidate inside a gap is evaluated with respect to both neighbors to ensure it connects well on both sides. For a gap $(L,R)$ with $L,R\in S$ and any $j\in(L,R)$, the joint edge score is the minimum of the boundary scores
\begin{equation}
J_{L,j,R}=\operatorname{min}\Bigl(T_{L,j}\cdot T_{j,R}\Bigr).   
\end{equation}
Subsequently, another sweep through the gaps is done, prioritizing larger temporal gaps, to select
\begin{equation}
 j^\star=\arg\max_{\,L<j<R} J_{L,j,R}   
\end{equation}
into $S$. Using the minimum filters out candidates that are weak to either neighbor, so each insertion bridges the gap with a frame that is simultaneously compatible with $L$ and $R$, improving coverage and local temporal consistency.

Conversely, when $|S|>K$, frames that add the least local value are removed using a neighbor-aware score. For each non-end index $t\in S$ with neighbors $L$ and $R$, we define
\begin{equation}
\Gamma_t \;=\; (T_{L,t} \;+\; T_{t,R})\; - \;T_{L,R}
\end{equation}
A small $\Gamma_t$ indicates that the direct connection $L\!\leftrightarrow\!R$ already carries strong utility while the two edges that rely on index $t$ are weak, making it a good candidate for removal. Finally, pruning proceeds by iteratively deleting $t^\star=\arg\min_t \Gamma_t$\, until $|S|=K$. This criterion preserves well-supported transitions, eliminates frames that do not improve local connectivity or viewpoint diversity, and yields a compact subset that maintains geometric coherence.

\vspace{1mm}
\noindent
\textbf{Implementation details.}
We implement BCVS using the basic Python-OpenCV library alone for fast CPU-only processing. Each image is resized (aspect preserved) to a maximum side of 960 pixels for feature extraction, and downsampled to 128 pixels for fast SSIM computation. ORB extracts up to $3000$ keypoints per frame; matches are computed with a BF–Hamming $k$-NN matcher and filtered by Lowe's ratio test $(\tau_{\text{ratio}}=0.75)$. The median inlier flow is normalized with a scale $\lambda=0.15$. Online selection uses a short temporal window $G=5$ and computes the global redundancy penalty $R^{G}_{j}$ over the last $L=20$ committed frames using an SSIM threshold $\tau=0.80$.

\newpage
\section{Benchmark Evaluation}

\begin{figure*}[t]
    \centering    \includegraphics[width=\textwidth]{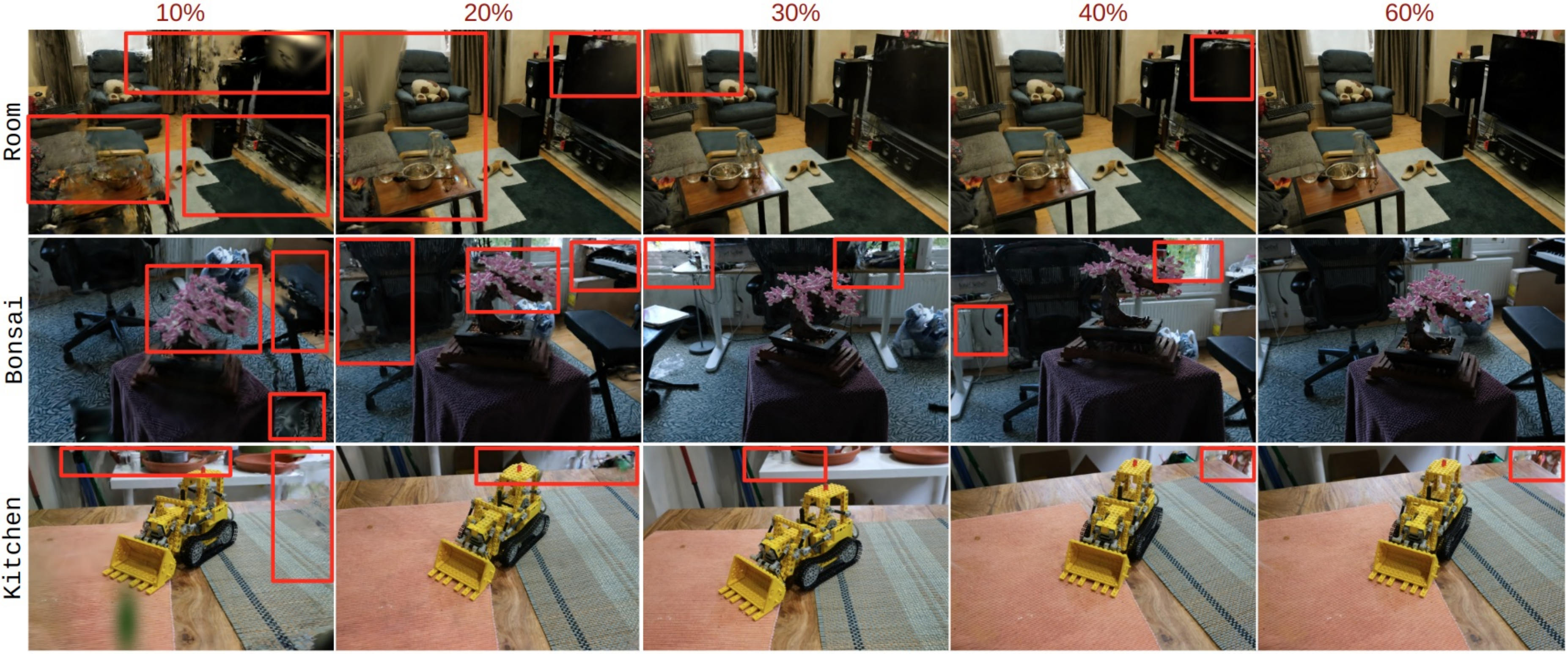}%
    \vspace{-1mm}
\caption{Additional qualitative evaluations of BCVS across sampling budgets (10–60\%) on MipNeRF-360~\cite{barron2022mip} data: across \texttt{room}, \texttt{kitchen}, and \texttt{bonsai} scenes. Similar to Fig.~\ref{Fig:qualitative}, the red boxes highlight major reconstruction errors and inconsistencies.}
    \label{Fig:qualitative_additional_scenes}
\end{figure*}

\subsection{Experimental Setup}
We assess the effectiveness of the proposed BCVS algorithm by training a 3D Gaussian Splatting (3DGS) model~\cite{kerbl20233d} on subsets of training views and measuring reconstruction quality on a held-out test set. To isolate the effect of view selection, all methods use the same 3DGS pipeline and identical hyperparameters. We sweep budgets of $10$-$90$\% with $10$\% intervals of the full data and report the qualitative and quantitative results.

\vspace{1mm}
\noindent\textbf{Dataset and Metrics.} We use the {MipNeRF-360}~\cite{barron2022mip} benchmark dataset, which provides forward-facing and $360^\circ$ trajectories with high-resolution images. According to standard practices, for each scene we construct a test set by assigning \textit{every $8$th frame}; all remaining frames form the training set. We train 3DGS for $30$k iterations per scene/budget using the default hyperparameters. Reconstruction quality is reported on the test split using PSNR and SSIM, averaged over test views and then across scenes. To ensure fairness, all runs share identical pre-processing, training resolution, and random seeds; only the training-view indices differ. In our evaluation, we use $7$ scenes: \texttt{\{bonsai, garden, kitchen, room, stump, bicycle, counter\}}; the per-scene image counts vary from {125} to {311} frames. 

\vspace{1mm}
\noindent \textbf{Performance Baselines.} We evaluate three baselines under the same protocol and budget $K$. The \textit{Random Sampler} picks $K$ training views randomly without replacement; we report mean and standard deviations over three seeds. The \textit{Uniform Sampler} takes one frame every $m$ frames along the capture order, where $m=\lfloor N/K\rfloor$ and $N$ is the number of frames; if we reach the end before selecting $K$ views, we continue from the beginning until we have exactly $K$, then keep them in chronological order. The \textit{Farthest-View Sampling (FVS)}~\cite{xiao2024nerf} greedily adds the view farthest from the current set in pose space, using the dataset’s camera intrinsics and extrinsics together with the sparse SfM point cloud to compute a photogrammetric distance, following the original NeRF Director formulation. Note that we consider view selection approaches that offer online CPU implementations for enabling deployments on resource-constrained mobile robots. Offline methods with GPU dependencies and prior pose estimation constraints are infeasible for real-time use. 

\subsection{Qualitative and Quantitative Analyses}
\begin{figure*}[t]
    \centering    \includegraphics[width=0.98\textwidth]{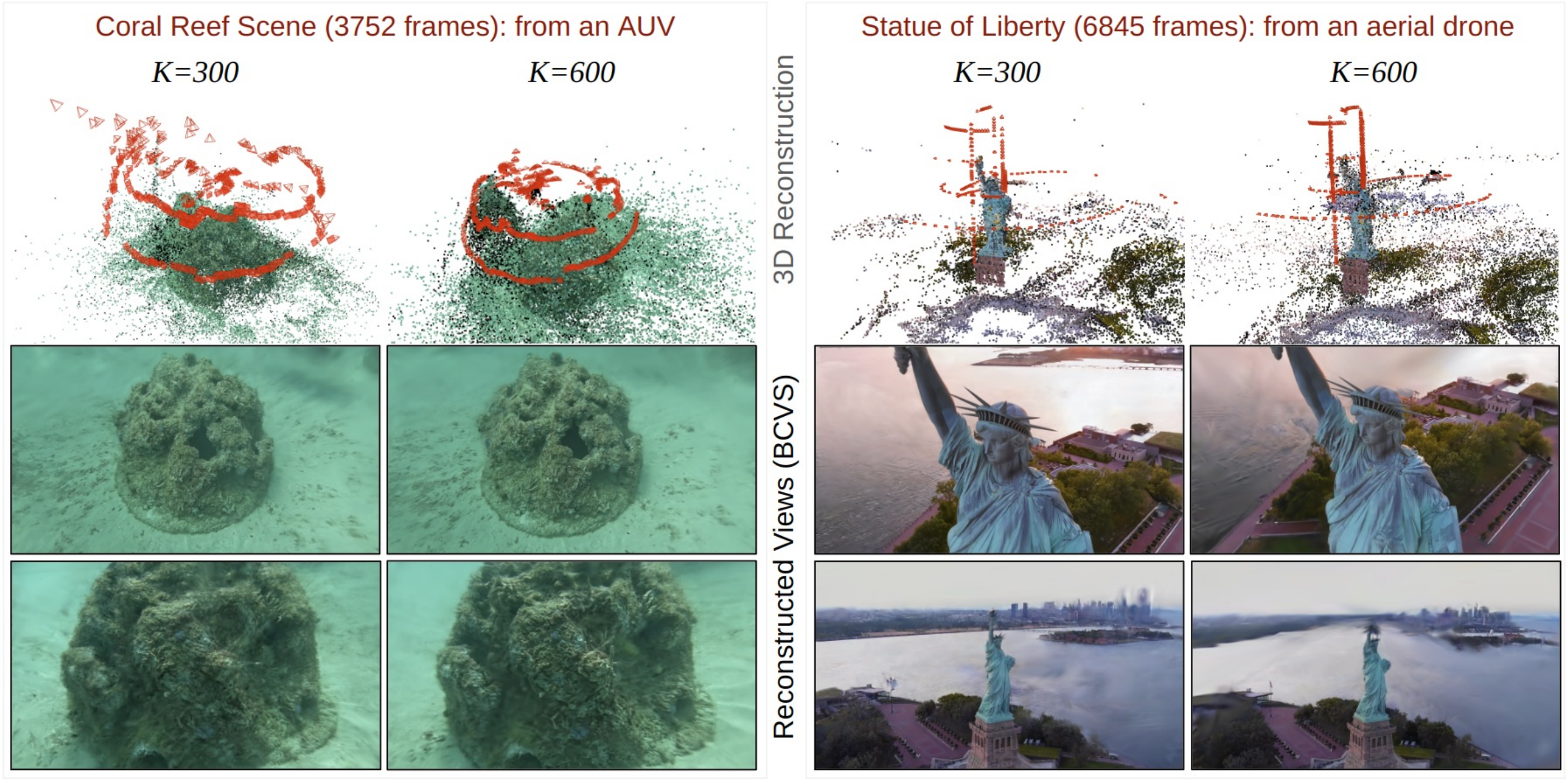}%
    \vspace{-1mm}
\caption{BCVS on long, noisy, real-world field trial data: a coral–reef survey data from an AUV on the left ($3752$ frames); and aerial views of the Statue of Liberty orbited by a drone on the right ($6845$ frames). We show sample results for two budgets ($K=300$ and $K=600$) for each scene. \textbf{Top rows:} COLMAP sparse reconstructions with selected camera poses (in red). \textbf{Bottom rows:} example rendered views from the trained 3DGS model. In the coral reef scene, we find that increasing the budget densifies coverage and reduces holes. On the other hand, in the aerial scene, increasing $K$ introduces many near-duplicate views with limited parallax, yielding noisier registration and blurrier renderings, demonstrating that adding frames without an added baseline can degrade reconstruction.}
\label{Fig:Field_data}
\vspace{-4mm}
\end{figure*}

Table \ref{results} presents the quantitative results for reconstruction quality across sampling budgets ($10$-$90$\%). To ensure fairness at low budgets, we exclude the \texttt{bicycle} and \texttt{stump} scenes from the $10$-$20$\% evaluations, as their limited number of images led to frequent COLMAP registration failures for most baselines. The comparative quantitative scores highlight three key insights. First, under tight budgets ($10$-$20$\%), BCVS provides the largest improvements, outperforming pose-based FVS by  $11.1$-$22.7$\% ($4.6$-$27.8$\%) on PSNR (SSIM). Although \textit{Random} and \textit{Uniform} samplers surpass FVS in this regime, they remain unstable across the scenes. This confirms that correspondence-guided novelty produces more informative and less redundant subsets when every frame matters. Second, in the mid-range of budget ($30$-$70$\%), all methods naturally improve as more views are available. \textit{Random} and \textit{Uniform} still vary across the scenes, while FVS typically averages higher than both; nevertheless, BCVS remains competitive to FVS without requiring poses. Finally, at high budgets ($80$-$90$\%), performance converges as expected, yet BCVS continues to match or surpass FVS. An additional practical advantage is runtime: BCVS operates entirely on CPU at $22.3$ FPS, enabling real-time processing without the pose-estimation overhead of FVS. 

Fig.~\ref{Fig:qualitative} illustrates qualitative results across sampling strategies evaluated on the \texttt{garden} scenes. Since each baseline yields a different subset, the per-subset test split produces different rendered images across methods. This fairly evaluates each baseline on viewpoints that are unseen relative to its own selection and reflects the coverage patterns induced by that baseline. Within the illustrations, red boxes highlight common reconstruction failures, including missing geometry, blurred textures, and ghosting artifacts~\cite{warburg2023nerfbusters}. 

Fig.~\ref{Fig:qualitative_additional_scenes} further presents additional results for BCVS on \texttt{room}, \texttt{kitchen}, and \texttt{bonsai} scenes at $10$-$60$\% budgets. We find that at low budgets ($10$-$20$\%), \textit{Random} and \textit{Uniform} samplers frequently leave wide gaps along the trajectory, producing tearing, holes, and incomplete structures. FVS distributes poses more broadly, sometimes improving sharpness but often yielding insufficient overlap between viewpoints, which leads to ghosting and localized distortions~\cite{warburg2023nerfbusters}. In contrast, BCVS consistently produces sharper boundaries, cleaner textures, and more complete geometry, even with less data. At mid to higher budgets ($30$–$70$\%), all methods improve, although \textit{Random} and \textit{Uniform} continue to exhibit residual inconsistencies, while FVS remains more reliable yet still prone to missing fine details when overlap is weak. Collectively, BCVS is the most robust, producing stable reconstructions with fewer artifacts. These qualitative observations align with the quantitative gains in Table~\ref{results}, reinforcing that correspondence-based novelty regularized by overlap delivers both higher accuracy and robustness.

\section{Field Experimental Evaluation}
\subsection{Field Trials Data: Aerial and Marine Robots}
We further evaluate BCVS on two long, noisy, real-world mission data collected by 6-DOF robots to test scalability and practical efficiency: (\textbf{i}) A drone orbit of the Statue of Liberty with $6,845$ frames; and (\textbf{ii}) An autonomous underwater vehicles (AUV) survey of a coral reef at the Atlantic, with $3,752$ frames. Both sequences contain long stretches of redundant images due to the slow motion and near-stationary viewpoints. To investigate the quality and efficiency trade-off, we run multiple budgets per mission, retaining $4.4$-$8.8$\% of the drone frames and $8$-$16$\% of the AUV frames, respectively. For these long streams of redundant data, we use a larger window gap ($G=20$), so that BCVS searches for the next distinct view over a wider temporal neighborhood, improving stability while keeping selection fast. We then train a standard 3DGS model on each selected subset, allowing us to examine how aggressively we can compress long missions for reconstruction, and how reconstruction quality and runtime evolve as $K$ increases.

Fig.~\ref{Fig:Field_data} compares visual results for two budgets on both mission data, as well as shows the COLMAP SfM outputs (camera frusta in red) and rendered viewpoints from 3DGS. On the coral reef sequence, increasing the budget from $300$ to $600$ raises viewpoint density along the ring trajectory, fills coverage gaps, and sharpens fine coral structure. Moreover, the reconstructed 3D point cloud is noticeably denser with fewer voids, an expected gain when added views contribute well-overlapped parallax. 

In contrast, on the aerial footage sequence, BCVS performs worse with $K=600$ than $K=300$ despite keeping more frames. We observed that many of the extra selections were near-duplicates along an almost circular, slow-motion path -- thus adding no new baselines while injecting repetitive correspondences. This eventually diluted the effective overlap and shortened the track lengths. Thus, the reconstructed point cloud was noisier with blurrier renderings.

\subsection{Discussion: Strengths and Weaknesses}
 BCVS is a pose-free and CPU-only algorithm, ideal for resource-constrained single-board devices. By regularizing novelty with overlap, it preferentially keeps parallax-rich, well-connected views and suppresses near-duplicates; as observed in our field data, the resulting subsets register robustly in COLMAP (most selected frames are incorporated), whereas naive FPS-based uniform sampling yields large baseline jumps or runs of similar frames that fail to register. However, BCVS is sensitive to the balance between connectivity and parallax: over-weighting connectivity risks keeping rotation-only redundancy, while prioritizing parallax can fragment overlap. Moreover, the feature matching quality can degrade in low-texture, dynamic, or heavily blurred scenes, which in turn weakens the selection signals. 
\section{Conclusion}
This report investigated a budget-aware view selection algorithm for efficient 3D reconstruction. We introduced BCVS, a real-time, pose-free view-selection pipeline that leverages keypoint correspondences to promote informative views and suppress redundancy. BCVS yields compact subsets that preserve geometric support with minimal reconstruction loss, running at over 22 FPS on CPU without pose estimation or deep encoders. Experiments on benchmark datasets as well as real-world aerial and marine robotics missions show that BCVS selects as little as 4–16\% of frames while sustaining clean registration and high-quality reconstructions. It outperforms CPU-friendly baselines at tight budgets by significant margins and remains competitive at higher budgets. Field experimental analyses further reveal that excessive near-duplicate views can degrade reconstructions, while budget-aware selection stabilizes geometry. These results demonstrate that simple correspondence cues are sufficient to achieve scalable, efficient, and accurate view selection -- ideal for use by mobile robots.


\vspace{-1mm}
\section*{Acknowledgment}
The authors acknowledge support from the Amazon Robotics grant AGR DTD 10-06-2025 and the National Science Foundation (NSF) award $2330416$. 


{
\bibliographystyle{IEEEtran}
\bibliography{refs,robopi_pubs}
}

\end{document}